\documentclass[runningheads]{llncs}
\usepackage{graphicx}
\usepackage{amsmath,amssymb} %
\usepackage{color}
\usepackage{times}
\usepackage{epsfig}
\usepackage{graphicx}
\usepackage{amsmath}
\usepackage{amssymb}
\usepackage{xspace}
\usepackage{array}
\usepackage{booktabs}
\usepackage{multirow}
\usepackage{etextools}
\usepackage{mathtools}
\usepackage{tabularx}
\usepackage{wrapfig,lipsum,booktabs}
\usepackage{floatrow}
\usepackage{subcaption}
\usepackage[font=small]{caption}
\usepackage[rightcaption]{sidecap}
\usepackage[table]{xcolor}

\newcommand{\setcapspacing}{\setlength\textfloatsep{.07in}}
\setcapspacing
\usepackage[pagebackref=true,breaklinks=true,letterpaper=true,colorlinks,bookmarks=false,hypertexnames=false]{hyperref}

\newcommand{\sect}[1]{Section~\ref{#1}}

\newcommand{\eqn}[1]{Equation~\ref{#1}}

\newcommand{\xsect}[1]{\vspace{-4.2mm}\section{#1}\vspace{-2.8mm}}
\newcommand{\xsub}[1]{\vspace{-4mm}\subsection{#1}\vspace{-1.8mm}}

\newcommand{\xpar}[1]{\vspace{-3.25mm}\paragraph{\normalfont\bf #1}\ \ }

\newcommand{\fig}[1]{Figure~\ref{#1}}
\newcommand{\tbl}[1]{Table~\ref{#1}}
\newcommand{\ignorethis}[1]{}
\newcommand{\norm}[1]{\lVert#1\rVert}

\newcommand{\ao}[1]{}

\newcommand{\aofoot}[1]{}

\newcolumntype{L}[1]{>{\raggedright\let\newline\\\arraybackslash\hspace{0pt}}m{#1}}
\newcolumntype{C}[1]{>{\centering\let\newline\\\arraybackslash\hspace{0pt}}m{#1}}
\newcolumntype{R}[1]{>{\raggedleft\let\newline\\\arraybackslash\hspace{0pt}}m{#1}}

\newcommand{\shiftdsetsize}{750,000\xspace}

\newcommand{\expec}[1]{\mathbb{E}_{\scriptscriptstyle{#1}}}

\makeatletter
\newcommand{\fixed@sra}{$\vrule height 2\fontdimen22\textfont2 width 0pt\shortrightarrow$}
\newcommand{\shortarrow}[1]{%
  \mathrel{\text{\rotatebox[origin=c]{\numexpr#1*45}{\fixed@sra}}}
}
\makeatother

\newcommand{\xho}{\hat{x}_1}
\newcommand{\xht}{\hat{x}_2}

\newcommand{\mc}[1]{\mathcal{#1}}

\newcommand{\eg}{{e.g.}\@\xspace}
\newcommand{\ie}{{i.e.}\@\xspace}
\newcommand{\etal}{{et al.}\@\xspace}

\makeatletter
\newcommand{\thickhline}{%
    \noalign {\ifnum 0=`}\fi \hrule height 1pt
    \futurelet \reserved@a \@xhline
}
\newcolumntype{"}{@{\hskip\tabcolsep\vrule width 1pt\hskip\tabcolsep}}
\makeatother

\newcommand{\projecturl}{\href{http://andrewowens.com/multisensory}{http://andrewowens.com/multisensory}}

\definecolor{lightyellow}{RGB}{255,255,170}
\newcommand{\bestcell}{\cellcolor{lightyellow}}

\begin{document}

\pagestyle{headings}
\mainmatter

\title{Audio-Visual Scene Analysis with\\Self-Supervised Multisensory Features}

\titlerunning{Audio-Visual Scene Analysis with Self-Supervised Multisensory Features}
\authorrunning{Owens and Efros}

\author{Andrew Owens
\quad
Alexei A. Efros}

\institute{UC Berkeley}

\maketitle

\vspace{-4.5mm}
\begin{abstract}

The thud of a bouncing ball, the onset of speech as lips open --- when visual and audio events occur together, it suggests that there might be a common, underlying event that produced both signals.
In this paper, we argue that the visual and audio components of a video signal should be modeled jointly using a fused multisensory representation.
We propose to learn such a representation in a self-supervised way, by training a neural network to predict whether video frames and audio are temporally aligned. We use this learned representation for three applications: (a) 
sound source localization, i.e. visualizing the source of sound in a video;
(b) audio-visual action recognition; and (c) on/off-screen audio source separation, e.g.
removing the off-screen translator's voice from a foreign official's speech.
Code, models, and video results are available on our webpage:~\projecturl.

\end{abstract}
\vspace{-3mm}
\xsect{Introduction}
\vspace{1mm}

As humans, we experience our world through a number of simultaneous
sensory streams. When we bite into an apple, not only do we taste it,
but --- as Smith and Gasser~\cite{smith2005development} point out ---
we also hear it crunch, see its red skin, and feel the coolness of its
core. The coincidence of sensations gives us strong evidence that they
were generated by a common, underlying event~\cite{sekuler1997sound},
since it is unlikely that they co-occurred across multiple modalities
merely by chance.  These cross-modal, temporal co-occurrences
therefore provide a useful learning signal: a model that is trained to
detect them ought to discover multi-modal structures that are useful
for other tasks.  In much of traditional computer vision research,
however, we have been avoiding the use of other, non-visual
modalities, arguably making the perception problem harder, not
easier.

In this paper, we learn a temporal, multisensory representation that fuses the visual and audio components of a video signal.
We propose to train this model without using any manually labeled data. 
That is, rather than explicitly telling the model that, \eg, it should associate moving lips with speech or a thud with a bouncing ball, we have it discover these audio-visual associations through self-supervised training~\cite{de1994learning}. 
Specifically, we train a neural network on a ``pretext'' task of detecting misalignment between audio and visual streams in synthetically-shifted videos. 
The network observes raw audio and video streams --- some of which are aligned, and some
that have been randomly shifted by a few seconds --- and we task it with distinguishing between the two. 
This turns out to be a challenging training task that forces the network to fuse visual motion with audio information and, in the process, learn a useful audio-visual feature representation.

\begin{figure}[t!]
  {\centering

\includegraphics[width=1.0\linewidth]{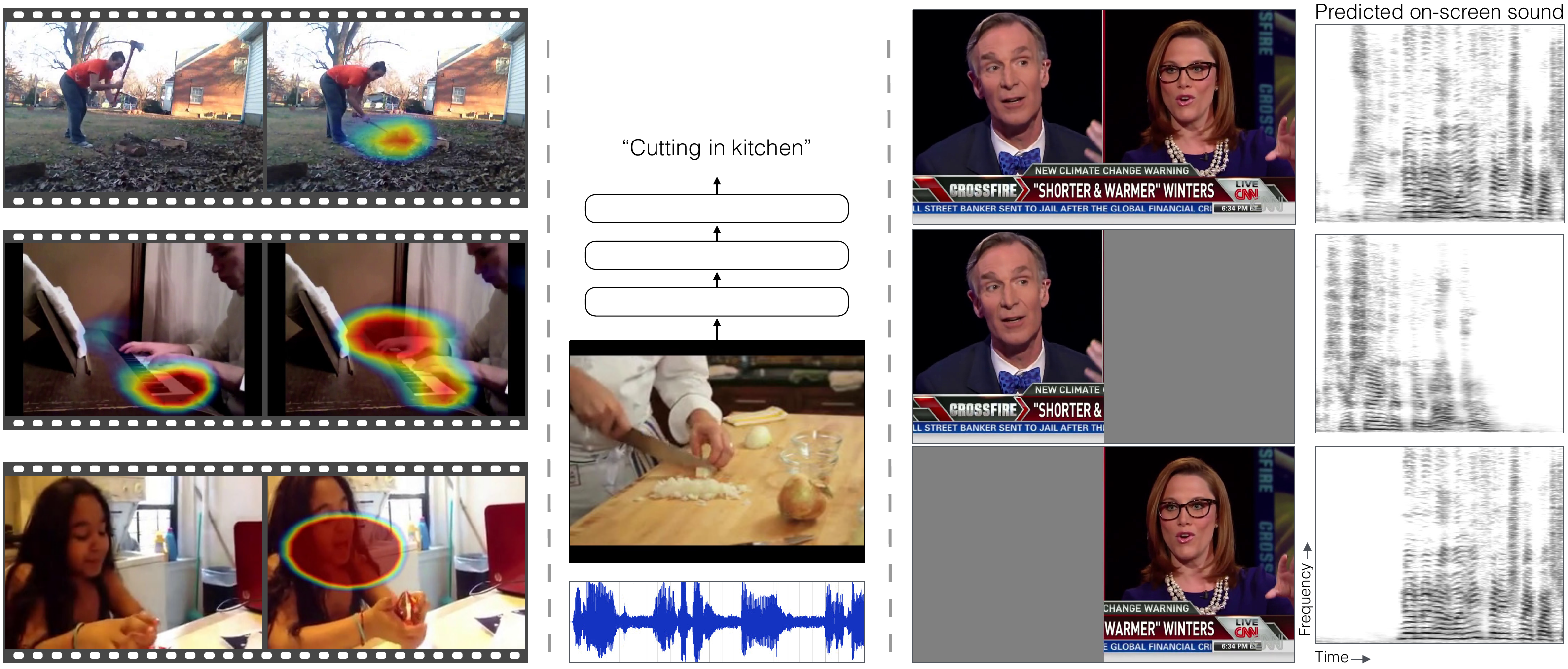}
  {\small  
  \begin{tabularx}{1.0\linewidth}{C{0.32\linewidth}C{0.25\linewidth}C{0.4\linewidth}}
    (a) Sound localization & (b) Action recognition & (c) On/off-screen audio separation
  \end{tabularx}}
  \caption{\small Applications. We use self-supervision to learn an audio-visual representation that: (a) can be used to visualize the
    locations of sound sources in video; (b) is useful for visual and
    audio-visual action recognition; (c) can be applied to the task of
    separating on- and off-screen sounds. In (c), we demonstrate our
    source-separation model by visually masking each speaker and
    asking it to predict the on-screen audio. The predicted sound
    contains only the voice of the visible speaker. Please see our
    webpage for video results:
    \projecturl.}\label{fig:teaser}}\vspace{-2mm}
\end{figure}

We demonstrate the usefulness of our multisensory representation in
three audio-visual applications: (a) sound source localization,
(b) audio-visual action recognition; and (c) on/off-screen sound source separation.  \fig{fig:teaser} shows examples of these applications.
In Fig. \ref{fig:teaser}(a), we visualize the sources of
sound in a video using our network's learned attention map, i.e. the impact of an axe, the opening of a mouth, and moving hands of a musician. In Fig. \ref{fig:teaser}(b), we show an application of our learned features to audio-visual action recognition, i.e. classifying a video of a chef chopping an
onion. In Fig. \ref{fig:teaser}(c), we demonstrate our novel on/off-screen sound source separation model's ability to separate the speakers' voices by visually masking them from the video. 

The main contributions of this paper are: 1) learning a general video
representation that fuses audio and visual information; 2) evaluating
the usefulness of this representation qualitatively (by sound source
visualization) and quantitatively (on an action recognition task); and
3) proposing a novel video-conditional source separation method that
uses our representation to separate on- and off-screen sounds, and is
the first method to work successfully on real-world video footage, \eg
television broadcasts.  Our feature representation, as well as code
and models for all applications are available online.

\vspace{-3.5mm}
\section{Related work}
\vspace{0.5mm}

\xpar{Evidence from psychophysics} While we often think of vision and
hearing as being distinct systems, in humans they are closely
intertwined~\cite{shimojo2001sensory} through a process known as {\em
  multisensory integration}. Perhaps the most compelling demonstration
of this phenomenon is the McGurk effect~\cite{mcgurk1976hearing}, an
illusion in which visual motion of a mouth changes one's
interpretation of a spoken sound\footnote{For a particularly vivid
  demonstration, please see:
  \url{https://www.youtube.com/watch?v=G-lN8vWm3m0} \cite{bbcseeing}.}.
Hearing can also influence vision: the timing of a sound, for instance,
affects whether we perceive two moving objects to be colliding or
overlapping~\cite{sekuler1997sound}. %
Moreover, psychologists have suggested that humans fuse audio and visual signals at a fairly early stage of processing~\cite{schwartz2002audio,omata2008fusion}, and that the two modalities are used jointly in perceptual grouping.
For example, the McGurk effect is less effective when the viewer first watches a video where audio and visuals in a video are unrelated, as this causes the signals to become ``unbound" (i.e. not grouped together) \cite{nahorna2012binding,nahorna2015audio}. This multi-modal perceptual grouping process is often referred to as {\em audio-visual scene analysis}~\cite{barker1998primitive,schwartz2002audio,hershey2004audio,nahorna2015audio}. 
In this paper, we take inspiration from psychology and propose a self-supervised multisensory feature representation as a computational model of audio-visual scene analysis.

\xpar{Self-supervised learning}  Self-supervised methods learn
features by training a model to solve a task derived from the
input data itself, without human labeling.  Starting with the early work of
de Sa~\cite{de1994learning}, there have been many self-supervised methods that learn to find correlations between sight and sound
~\cite{ngiam2011multimodal,owens2015visually,owens2016ambient,arandjelovic2017look}. These methods, however, have either learned the correspondence between
static images and ambient sound~\cite{owens2016ambient,arandjelovic2017look}, or have analyzed motion
in very limited domains~\cite{owens2015visually,ngiam2011multimodal}
(\eg \cite{owens2015visually} only modeled drumstick impacts). Our learning task resembles
Arandjelovi{\'c} and Zisserman~\cite{arandjelovic2017look}, which predicts whether an image and an
audio track are sampled from the same (or different) videos. Their task, however, is solvable
from a single frame by recognizing semantics (\eg indoor vs. outdoor scenes). Our inputs, by contrast, always come from the same video, and we predict whether they are aligned; hence our task requires motion analysis to solve.
Time has also been used as supervisory signal, e.g. predicting the temporal ordering in a video \cite{misra2016shuffle,wei2018arrow,fernando2017self}. In contrast, our network learns to analyze audio-visual actions, which are likely to correspond to salient physical processes.

\xpar{Audio-visual alignment} While we study alignment for
self-supervised learning, it has also been studied as an end in
itself~\cite{mcallister1997lip,marcheret2015detecting,chung2016out} \eg
in lip-reading applications~\cite{chung2017lip}.  Chung and
Zisserman~\cite{chung2016out}, the most closely related approach,
train a two-stream network with an embedding loss. Since aligning
speech videos is their end goal, they use a face detector (trained
with labels) and a tracking system to crop the speaker's
face. This allows them to address the problem with a 2D CNN that
takes 5 channel-wise concatenated frames cropped around a mouth as
input (they also propose using their image features for
self-supervision; while promising, these results are very
preliminary).

\xpar{Sound localization} The goal of visually locating the source of sounds in a video has a long history.  The seminal work of Hershey \etal \cite{hershey1999audio}
localized sound sources by measuring mutual information between visual
motion and audio using a Gaussian process model.  Subsequent work also considered
subspace methods \cite{fisher2000learning}, canonical correlations
\cite{kidron2005pixels}, and keypoints \cite{barzelay2007harmony}. Our model learns to associate motions with sounds via
self-supervision, without us having to explicitly model them.

\xpar{Audio-Visual Source Separation} Blind source separation (BSS), i.e. separating the individual sound sources in an audio stream --- also known as the {\em cocktail party}
problem \cite{cherry1953some} --- is a classic audio-understanding task~\cite{bregman1994auditory}.  Researchers have proposed many successful probabilistic approaches to this problem
\cite{ghahramani1996factorial,roweis2001one,cooke2010monaural,virtanen2007monaural}. More recent deep learning approaches involve predicting an
embedding that encodes the audio clustering~\cite{hershey2016deep,chen2017attractor}, or optimizing a 
  permutation invariant loss~\cite{yu2017permutation}. 
It is natural to also want to include the visual signal to solve this problem, often referred to as {\em Audio-Visual Source Separation}. For example,
\cite{darrell2000audio,fisher2000learning} masked frequencies based on
their correlation with optical flow; \cite{hershey2004audio} used graphical models; \cite{barzelay2007harmony} used priors on harmonics;
\cite{pu2017audio} used a sparsity-based factorization method; and
\cite{casanovas2010blind} used a clustering method. Other methods use face detection and
multi-microphone beamforming~\cite{rivet2014audiovisual}. These methods make strong
assumptions about the relationship between sound and motion, and have
mostly been applied to lab-recorded video. Researchers have proposed learning-based methods that address these limitations, e.g. 
~\cite{khan2013speaker} use mixture models to predict separation masks. Recently,~\cite{hou2017audio} proposed a convolutional network that isolates on-screen speech, although this model is relatively small-scale (tested on videos from one
speaker). We do on/off-screen source separation on
more challenging internet and broadcast videos by combining our representation with a
$u$-net~\cite{ronneberger2015u} regression model.

\xpar{Concurrent work}
Concurrently and independently from us, a number of groups have
proposed closely related methods for source separation and sound
localization. Gabbay \etal~\cite{gabbay2017seeing,gabbay2017visual}
use a vision-to-sound method to separate speech, and propose a
convolutional separation model. Unlike our work, they
assume speaker identities are
known. Ephrat \etal~\cite{ephrat2018looking} and
Afouras \etal~\cite{afouras2018conversation} 
separate the speech of a user-chosen speaker from videos containing
multiple speakers, using face detection and tracking systems
to group the different speakers. Work by
Zhao \etal~\cite{zhao2018sound} and Gao \etal~\cite{gao2018learning}
separate sound for multiple visible objects (\eg musical
instruments). This task involves associating objects with the sounds
they typically make based on their appearance, while ours involves the
``fine-grained'' motion-analysis task of separating multiple speakers. There has also been recent work on localizing sound sources
using a network's attention map
\cite{senocak2018learning,arandjelovic2017objects,owens2017learning}.
These methods are similar to ours, but they largely localize objects
and ambient sound in static images, while ours responds to actions in videos.

\xsect{Learning a self-supervised multisensory representation}
\label{sec:rep}
\begin{figure}[t!]
  {\centering
        \includegraphics[width=0.85\linewidth]{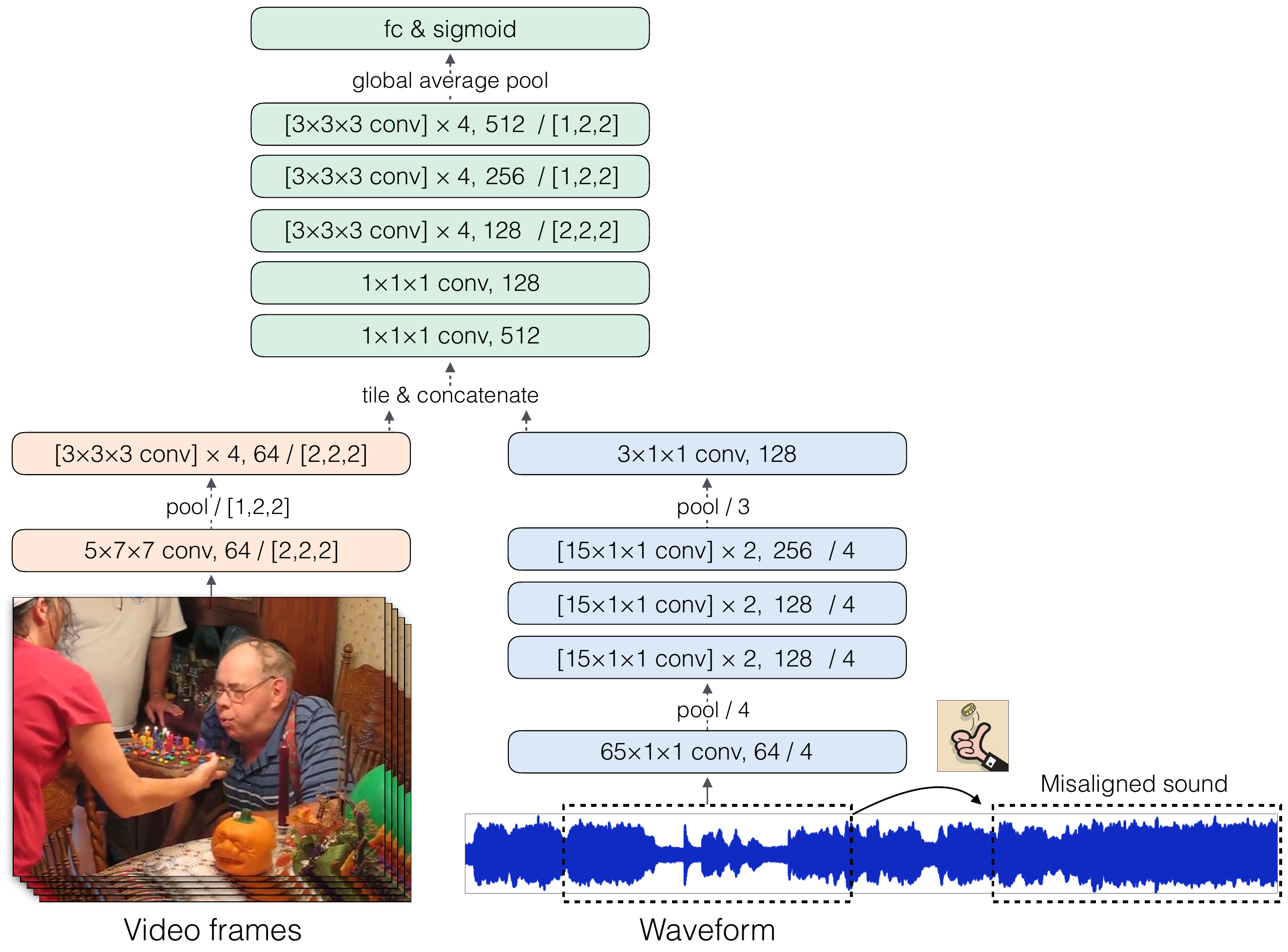}} %
  \vspace{-2mm}
    \caption{\small Fused audio-visual network. We train an early-fusion,
      multisensory network to predict whether video frames and audio
      are temporally aligned. We include residual connections between
      pairs of convolutions \cite{he2016deep}. We represent the input
      as a $T \times H \times W$ volume, and denote a stride by
      ``/2''. To generate misaligned samples, we synthetically shift
      the audio by a few seconds.}
  \label{fig:network}
\end{figure}

We propose to learn a representation using self-supervision, by
training a model to predict whether a video's audio and visual streams
are temporally synchronized.

\xpar{Aligning sight with sound} During training, we feed a neural
network video clips. In half of them, the vision and sound streams are
synchronized; in the others, we shift the audio by a few seconds. We
train a network to distinguish between these examples. More
specifically, we learn a model $p_\theta(y \mid I, A)$ that predicts
whether the image stream $I$ and audio stream $A$ are synchronized, by
maximizing the log-likelihood:
\vspace{-0.4mm}
\begin{equation}
  \mathcal{L}(\theta) = \frac{1}{2} \expec{I, A, t}[\log(p_\theta(y =
  1 \mid I, A_0)) + \log(p_\theta(y = 0 \mid I, A_t))],
\end{equation}
\vspace{-0.3mm}
where $A_s$ is the audio track shifted by $s$ secs., $t$ is a random
temporal shift, $\theta$ are the model parameters, and $y$ is the
event that the streams are synchronized.  This learning problem is
similar to noise-contrastive estimation~\cite{gutmann2010noise}, which
trains a model to distinguish between real examples and noise; here,
the noisy examples are misaligned videos.

\xpar{Fused audio-visual network design} Solving this task requires
the integration of low-level information across modalities. In order
to detect misalignment in a video of human speech, for instance, the
model must associate the subtle motion of lips with the timing of
utterances in the sound.  We hypothesize that early fusion of audio
and visual streams is important for modeling actions that produce a
signal in both modalities. We therefore propose to solve our task
using a 3D multisensory convolutional network (CNN) with an
early-fusion design (\fig{fig:network}).

Before fusion, we apply a small number of 3D convolution and pooling
operations to the video stream, reducing its temporal sampling rate by
a factor of 4. We also apply a series of strided 1D convolutions to
the input waveform, until its sampling rate matches that of the video
network. We fuse the two subnetworks by concatenating their
activations channel-wise, after spatially tiling the audio
activations.  The fused network then undergoes a series of 3D
convolutions, followed by global average
pooling~\cite{lin2013network}. We add residual connections between
pairs of convolutions. We note that the network architecture resembles
ResNet-18~\cite{he2016deep} but with the extra audio subnetwork, and
3D convolutions instead of 2D ones (following work on inflated
convolutions \cite{carreira2017quo}).

\xpar{Training} We train our model with 4.2-sec. videos, 
randomly shifting the audio by 2.0 to 5.8 seconds. We train our model
on a dataset of approximately \shiftdsetsize videos randomly sampled
from AudioSet~\cite{gemmeke2017audio}. We use full frame-rate videos
(29.97 Hz), resulting in 125 frames per example. We select random $224
\times 224$ crops from resized $256 \times 256$ video frames, apply
random left-right flipping, and use 21 kHz stereo sound. We
sample these video clips from longer (10 sec.) videos.
Optimization details can be found in \sect{sec:opt}. %

\xpar{Task performance} We found that the model obtained 59.9\%
accuracy on held-out videos for its alignment task (chance $= 50\%$).
While at first glance this may seem low, we note that in many videos
the sounds occur off-screen~\cite{owens2016ambient}. Moreover, we
found that this task is also challenging for humans.
To get a better understanding of human ability, we showed 30
participants from Amazon Mechanical Turk 60 aligned/shifted video
pairs, and asked them to identify the one with out-of-sync sound. We
gave them 15 secs. of video (so they have significant temporal
context) and used large, 5-sec. shifts. They solved the task
with $66.6\% \pm 2.4\%$ accuracy.

To help understand what actions the model can predict synchronization
for, we also evaluated its accuracy on categories from the Kinetics
dataset~\cite{kay2017kinetics}
(\fig{fig:kineticsacc}). It was most successful for
classes involving human speech: \eg, {\em news anchoring}, {\em
  answering questions}, and {\em testifying}. Of course, the most
important question is whether the learned audio-visual representation
is useful for downstream tasks. We therefore turn out attention to
applications.

\begin{figure}[t!]
  \centering
      {\includegraphics[width=1.0\linewidth]{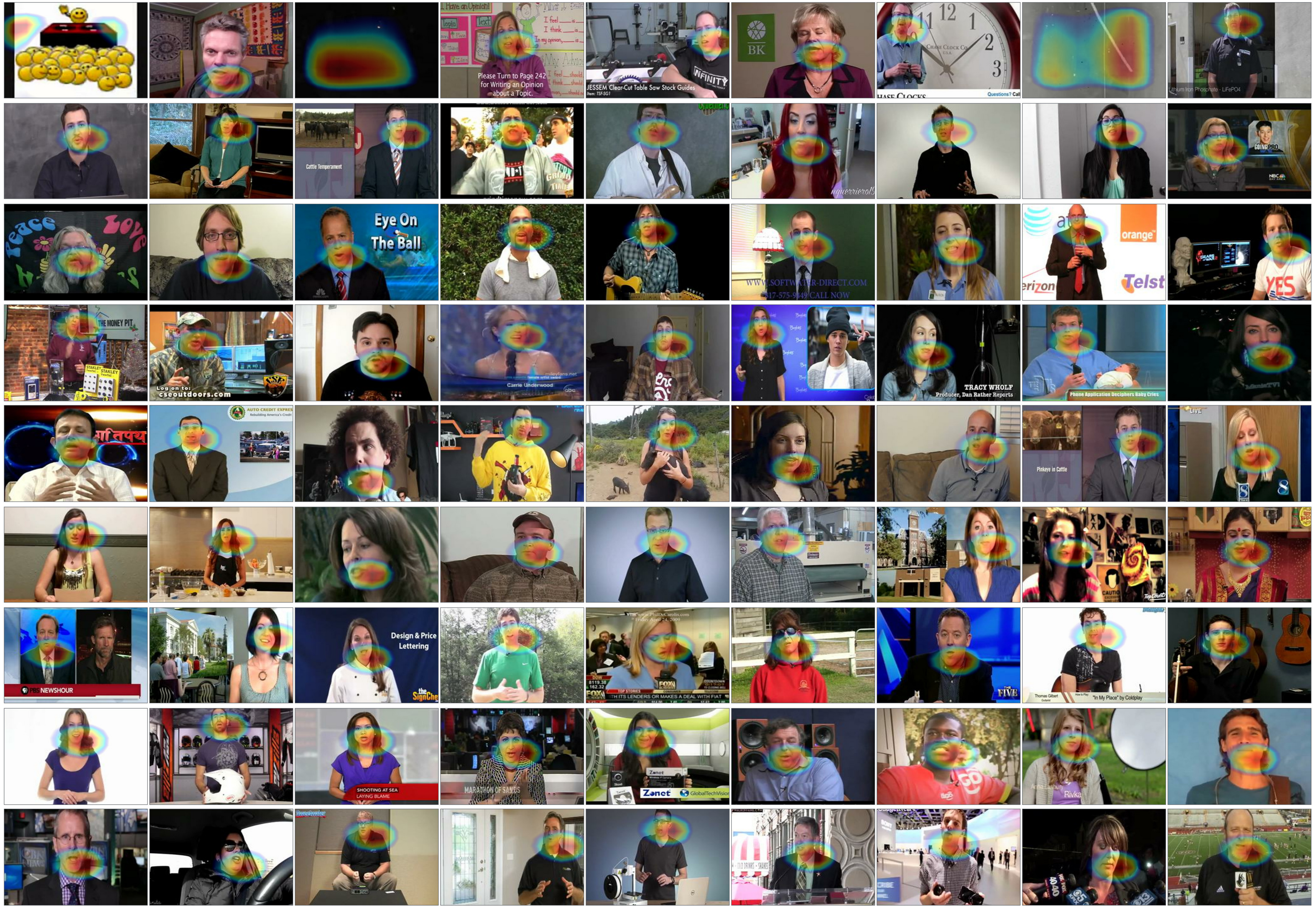}}
  \caption{{\small Visualizing sound sources. We show the video frames in held-out AudioSet
      videos with the strongest class activation map (CAM) response (we scale its range per
      image to compensate for the wide range of values).}}
    \label{fig:audiosetlocalize}\vspace{-3.2mm}
\end{figure}
\begin{figure}
  \vspace{-4mm}
  \centering
  \includegraphics[width=\linewidth]{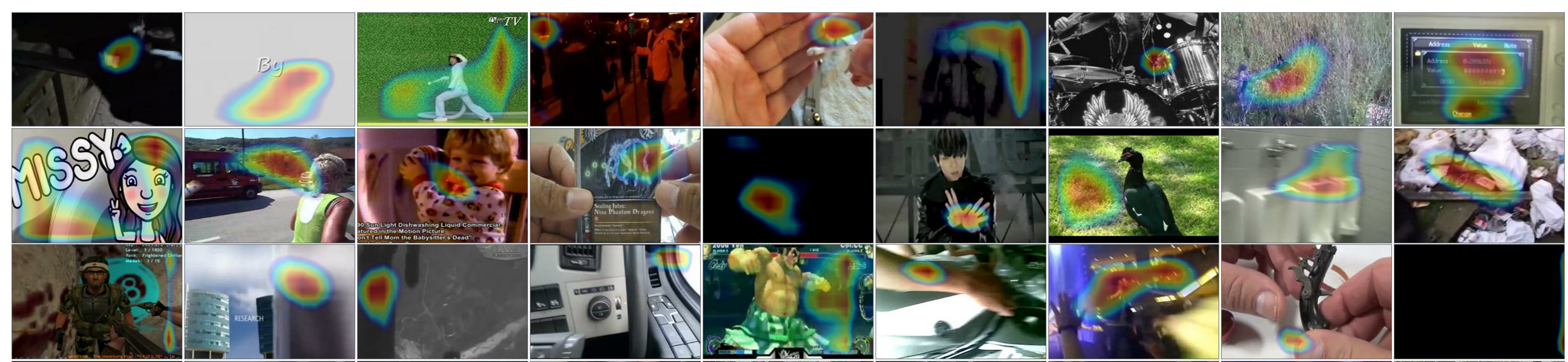}
  \caption{{\small Examples with the weakest class activation map
      response (c.f. \fig{fig:audiosetlocalize}).}}\vspace{-3mm}
  \label{fig:uncertain}
\end{figure}

\begin{figure}[ht!]
  \centering
      {\includegraphics[width=1.0\linewidth]{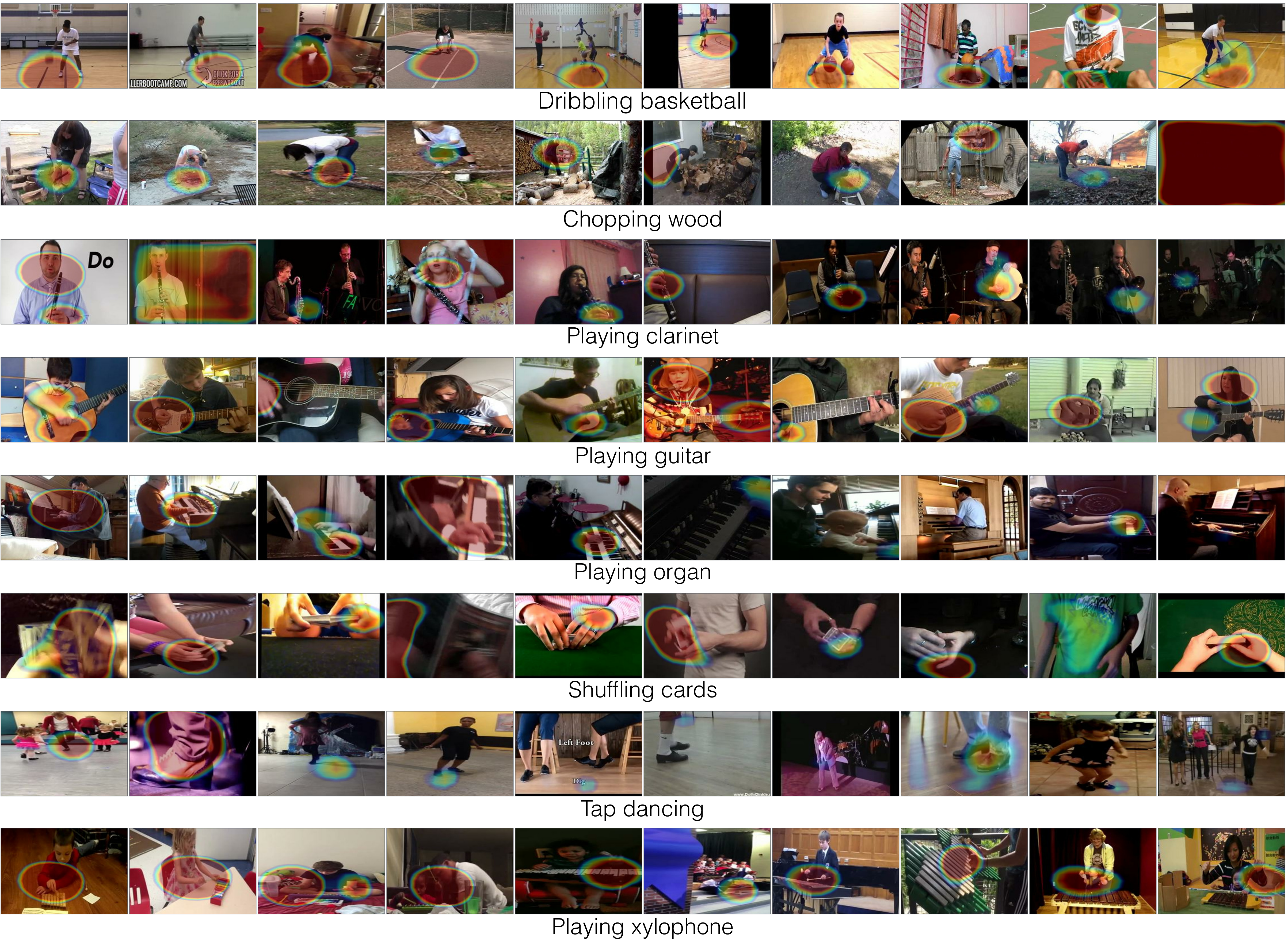}}
  \caption{{\small Strongest CAM responses for classes in the
      Kinetics-Sounds dataset \cite{arandjelovic2017look}, after
      manually removing frames in which the activation was only to a
      face (which appear in almost all categories). We note that no
      labeled data was used for training. We do not rescale the heat
      maps per image (\ie the range used in this visualization is
      consistent across examples). }}\vspace{-2.9mm}
  \label{fig:kineticslocalize}
\end{figure}

\xsect{Visualizing the locations of sound sources}
\label{sec:visualize}

One way of evaluating our representation is to visualize the
audio-visual structures that it detects. A good audio-visual
representation, we hypothesize, will pay special attention to {\em
  visual sound sources} --- on-screen actions that make a sound, or
whose motion is highly correlated with the onset of sound.  We note
that there is ambiguity in the notion of a sound source for
in-the-wild videos. For example, a musician's lips, their larynx, and
their tuba could all potentially be called the source of a sound. Hence
we use this term to refer to motions that are correlated with
production of a sound, and study it through network visualizations.

To do this, we apply the class activation map (CAM) method of Zhou
\etal~\cite{zhou2016learning}, which has been used for localizing
ambient sounds~\cite{owens2017learning}. Given a space-time video
patch $I_x$, its corresponding audio $A_x$, and the features assigned
to them by the last convolutional layer of our model, $f(I_x, A_x)$,
we can estimate the probability of alignment with:
\vspace{-0.5mm}
\begin{equation}
  p(y \mid I_x, A_x) = \sigma (w^\top f(I_x, A_x)),
  \label{eq:patchprob}
\end{equation}
where $y$ is the binary alignment label, $\sigma$ the
sigmoid function, and $w$ is the model's final affine layer. We can
therefore measure the information content of a patch --- and, by our
hypothesis, the likelihood that it is a sound source --- by the
magnitude of the prediction $|w^\top f(I_x, A_x)|$.

One might ask how this self-supervised approach to localization
relates to generative approaches, such as classic mutual information
methods \cite{hershey1999audio,fisher2000learning}. To help understand
this, we can view our audio-visual observations as having been
produced by a generative process (using an analysis similar
to~\cite{isola2015learning}): we sample the label $y$, which
determines the alignment, and then conditionally sample $I_x$ and
$A_x$.  Rather than computing mutual information between the two
modalities (which requires a generative model that self-supervised
approaches do not have), we find the patch/sound that provides the
most information about the latent variable $y$, based on our learned
model $p(y \mid I_x, A_x)$. %

\xpar{Visualizations} What actions does our network respond to? First,
we asked which space-time patches in our test set were most
informative, according to \eqn{eq:patchprob}. We show the top-ranked
patches in \fig{fig:audiosetlocalize}, with the class activation map
displayed as a heatmap and overlaid on its corresponding video frame.
From this visualization, we can see that the network is selective to
faces and moving mouths.  The strongest responses that are not faces
tend to be unusual but salient audio-visual stimuli (\eg two
top-ranking videos contain strobe lights and music). For comparison,
we show the videos with the weakest response in \fig{fig:uncertain};
these contain relatively few faces.

Next, we asked how the model responds to videos that do not contain
speech, and applied our method to the Kinetics-Sounds
dataset~\cite{arandjelovic2017look} --- a subset of
Kinetics~\cite{kay2017kinetics} classes that tend to contain a
distinctive sound. We show the examples with the highest response for
a variety of categories, after removing examples in which the response
was solely to a face (which appear in almost every category). We show
results in \fig{fig:kineticslocalize}.

Finally, we asked how the model's attention varies with motion. To
study this, we computed our CAM-based visualizations for videos, which
we have included in the supplementary video (we also show some
hand-chosen examples in \fig{fig:teaser}(a)). These results
qualitatively suggest that the model's attention varies with on-screen
motion. This is in contrast to single-frame methods
models~\cite{senocak2018learning,owens2017learning,arandjelovic2017look},
which largely attend to sound-making objects rather than actions.

\vspace{-2mm}
\xsect{Action recognition}
\label{sec:action}

{\setlength\textfloatsep{10mm}
\begin{SCtable}[\sidecaptionrelwidth][t!]
{{\small
  \setlength{\aboverulesep}{0pt}
  \setlength{\belowrulesep}{0pt}

\begin{minipage}{0.5\linewidth}
    \begin{tabularx}{1.0\linewidth}{L{0.85\linewidth}|R{0.15\linewidth}}

    Model & Acc.~~~ \\
       \toprule
    Multisensory (full) & {\bestcell 82.1\%} \\
    Multisensory (spectrogram) & 81.1\% \\
    Multisensory (random pairing~\cite{arandjelovic2017look}) & 78.7\% \\   
    Multisensory (vision only) & 77.6\% \\
    Multisensory (scratch) & 68.1\% \\
    I3D-RGB (scratch) \cite{carreira2017quo} & 68.1\% \\
    O3N \cite{fernando2017self}* & 60.3\% \\
    Purushwalkam \etal \cite{purushwalkam2016pose}* & 55.4\% \\
    C3D \cite{tran2014c3d,carreira2017quo}* & 51.6\% \\ 
    Shuffle \cite{misra2016shuffle}* & 50.9\% \\
    Wang \etal \cite{wang2015unsupervised,purushwalkam2016pose}* & 41.5\% \\
    \hline
    I3D-RGB + ImageNet \cite{carreira2017quo} & 84.2\% \\
    I3D-RGB + ImageNet + Kinetics \cite{carreira2017quo} & {\bestcell 94.5\%} \\

    \end{tabularx}%
    \vspace{5mm}
  \end{minipage}}\hspace{-3.5mm} %
  \caption{\small Action recognition on UCF-101 (split 1). We compared
    methods pretrained without labels (top), and with semantic labels
    (bottom). Our model, trained both with and without sound,
    significantly outperforms other self-supervised methods. Numbers
    annotated with ``*'' were obtained from their corresponding
    publications; we retrained/evaluated the other models.} \label{tab:ucff}\vspace{-3mm}
}
\vspace{-3.2mm}
\end{SCtable}}
\setlength\textfloatsep{.09in}

We have seen through visualizations that our representation conveys
information about sound sources.  We now ask whether it is useful for
recognition tasks.  To study this, we fine-tuned our model for
action recognition using the UCF-101 dataset~\cite{soomro2012ucf101},
initializing the weights with those learned from our alignment task.  We
provide the results in \tbl{tab:ucff}, and compare our model to 
other unsupervised learning and 3D CNN methods.

We train with 2.56-second subsequences, following
\cite{carreira2017quo}, which we augment with random flipping and
cropping, and small (up to one frame) audio shifts. At test time, we
follow~\cite{simonyan2014two} and average the model's outputs over 25
clips from each video, and use a center $224 \times 224$ crop.
Please see \sect{sec:opt} for optimization details. %

\xpar{Analysis} We see, first, that our model significantly
outperforms self-supervised approaches that have previously been
applied to this task, including Shuffle-and-Learn
\cite{misra2016shuffle} (82.1\% vs. 50.9\% accuracy) and O3N
\cite{fernando2017self} (60.3\%). We suspect this is in part due to
the fact that these methods either process a single frame or a short
sequence, and they solve tasks that do not require extensive motion
analysis. We then compared our model to methods that use supervised
pretraining, focusing on the state-of-the-art I3D
\cite{carreira2017quo} model. While there is a large gap between our
self-supervised model and a version of I3D that has been pretrained on
the closely-related Kinetics dataset (94.5\%), the performance of our
model (with both sound and vision) is close to the (visual-only) I3D
pretrained with ImageNet~\cite{deng2009imagenet} (84.2\%).

Next, we trained our multisensory network with the self-supervision
task of~\cite{arandjelovic2017look} rather than our own, i.e. creating
negative examples by randomly pairing the audio and visual streams
from different videos, rather than by introducing misalignment. We
found that this model performed significantly worse than ours
(78.7\%), perhaps due to the fact that its task can largely be solved
without analyzing motion.

Finally, we asked how components of our model contribute to its
performance. To test whether the model is obtaining its predictive
power from audio, we trained a variation of the model in which the
audio subnetwork was ablated (activations set to zero), finding that
this results in a 5\% drop in performance. This suggests both that
sound is important for our results, and that our visual features are
useful in isolation. We also tried training a variation of the model
that operated on spectrograms, rather than raw waveforms, finding that
this yielded similar performance (\sect{sec:appbaseline}). To measure
the importance of our self-supervised pretraining, we compared our
model to a randomly initialized network (\ie trained from scratch),
finding that there was a significant (14\%) drop in performance ---
similar in magnitude to removing ImageNet pretraining from I3D. These
results suggest that the model has learned a representation that is
useful both for vision-only and audio-visual action recognition.

\xsect{On/off-screen audio-visual source separation}
\begin{figure}[t!]
  \centering
  \includegraphics[width=0.63\linewidth]{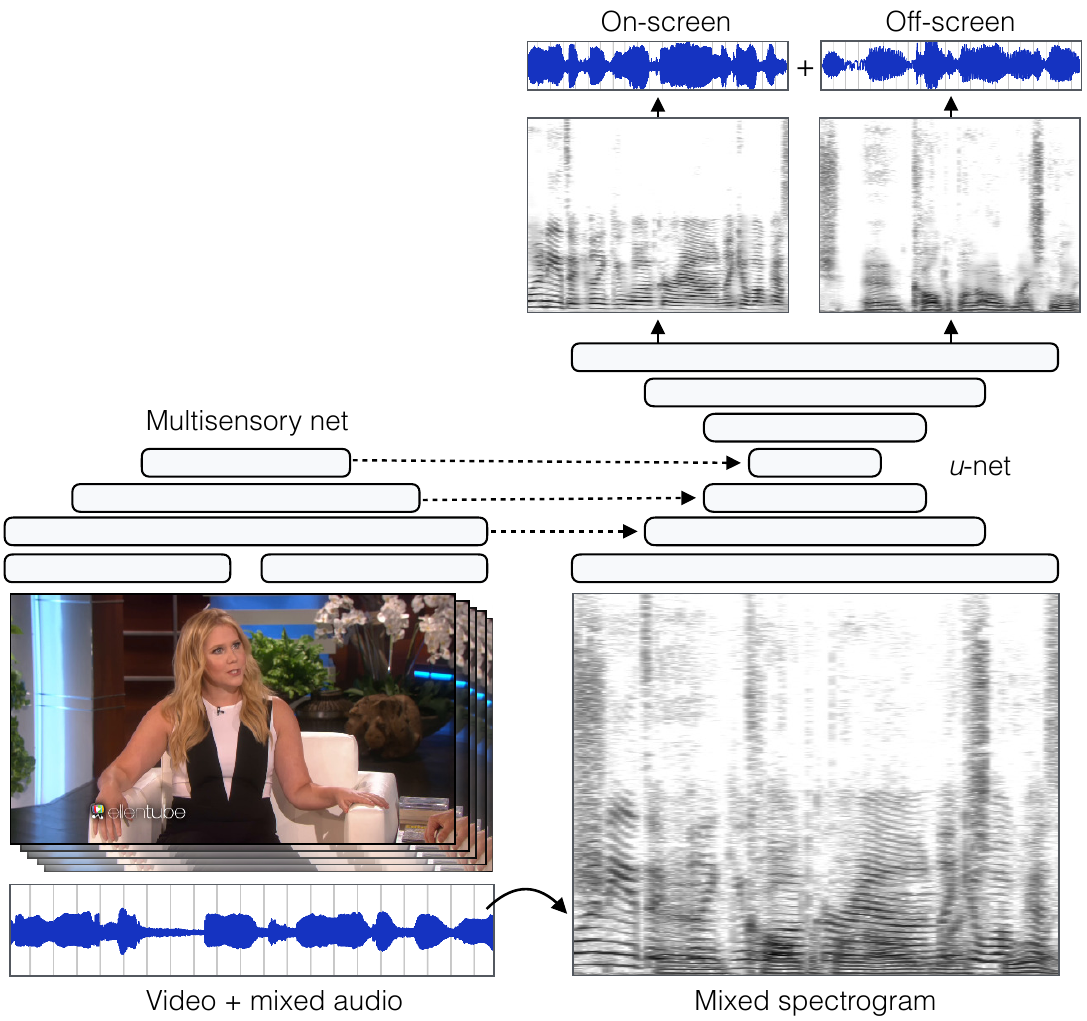} %
  \vspace{-2.5mm}
  \caption{\small Adapting our audio-visual network to a source
    separation task. Our model separates an input spectrogram into on-
    and off-screen audio streams. After each temporal downsampling
    layer, our multisensory features are concatenated with those of a
    $u$-net computed over spectrograms. We invert the spectrograms to
    obtain waveforms. The model operates on raw video, without any
    preprocessing (\eg no face detection).}
  \label{fig:sepnet}
\end{figure}

We now apply our representation to a classic audio-visual
understanding task: separating on- and off-screen sound. To do this,
we propose a source separation model that uses our learned
features. Our formulation of the problem resembles recent audio-visual
and audio-only separation work
\cite{hershey2016deep,yu2017permutation,huang2015separation,hou2017audio}.
We create synthetic sound mixtures by summing an input video's
(``on-screen'') audio track with a randomly chosen (``off-screen'')
track from a random video. Our model is then tasked with separating
these sounds.  %

\xpar{Task} We consider models that take a spectrogram for the mixed
audio as input and recover spectrogram for the two mixture
components. Our simplest on/off-screen separation model learns to
minimize:
\vspace{-0.25mm}
\begin{equation}
  \mc{L_O}(x_M, I) =  \norm{x_F - f_F(x_M, I)}_1 + \norm{x_B - f_B(x_M, I)}_1,
  \label{eq:onoffloss}
\end{equation}
\vspace{-0.25mm}
where $x_M$ is the mixture sound, $x_F$ and $x_B$ are the spectrograms
of the on- and off-screen sounds that comprise it (\ie foreground and
background), and $f_F$ and $f_B$ are our model's predictions of them
conditional on the (audio-visual) video $I$.

We also consider models that segment the two sounds without regard for
their on- or off-screen provenance, using the permutation invariant
loss (PIT) of Yu \etal \cite{yu2017permutation}. This loss is similar
to \eqn{eq:onoffloss}, but it allows for the on- and off-screen sounds
to be swapped without penalty:
\vspace{-0.4mm}
\begin{equation}
  \mc{L_P}(x_F, x_B, \xho, \xht) = \min(L(\xho, \xht), L(\xht, \xho)),
  \label{eq:pit}
\end{equation}
\vspace{-0.3mm}
where $L(x_i, x_j) = \norm{x_i - x_F}_1 + \norm{x_j - x_B}_1$ and
$\xho$ and $\xht$ are the predictions.

\xsub{Source separation model}
\setlength{\tabcolsep}{2.5pt}

\begin{table}[t!]
  {\centering \small
    \begin{tabularx}{\linewidth}{l C{.07\linewidth}
    C{.06\linewidth} C{.06\linewidth} C{.06\linewidth}
    C{.07\linewidth} C{.06\linewidth}
    C{.07\linewidth} C{.06\linewidth}
    C{.07\linewidth} C{.06\linewidth}}
  \toprule
    \multirow{1}{*}{Method}
  & \multicolumn{4}{c}{All}
  & \multicolumn{2}{c}{Mixed sex}
  & \multicolumn{2}{c}{Same sex}
  & \multicolumn{2}{c}{GRID transfer} \\
  \cmidrule(r{2pt}){2-5}  
  \cmidrule(l{2pt}){6-7}
  \cmidrule(l{2pt}){8-9}
  \cmidrule(l{2pt}){10-11}
  & On/off & SDR & SIR & SAR
  & On/off & SDR
  & On/off & SDR
  & On/off & SDR \\
  On/off + PIT & {\bestcell 11.2} & {\bestcell 7.6} & {\bestcell 12.1} & 10.2    & {\bestcell 10.6} & {\bestcell 8.8} &    {\bestcell 11.8} & {\bestcell 6.5} & {\bestcell 13.0} & 7.8 \\
  Full on/off  & 11.4 & 7.0 & 11.5 & 9.8                       & 10.7 & 8.4             & 11.9 & 5.7 & 13.1 & 7.3 \\ 
  Mono         & 11.4 & 6.9 & 11.4 & 9.8                       & 10.8 & 8.4             & 11.9 & 5.7 & 13.1 & 7.3   \\
  Single frame & 14.8 & 5.0 & 7.8 & {\bestcell 10.3}                 & 13.2 & 7.2             & 16.2 & 3.1 & 17.8 & 5.7   \\
  No early fusion & 11.6 & 7.0 & 11.0 & 10.1                     & 11.0 & 8.4             & 12.1 & 5.7 & 13.5 & 6.9 \\
  Scratch      & 12.9 & 5.8 & 9.7 & 9.4                       & 11.8 & 7.6              & 13.9 & 4.2 & 15.2 & 6.3   \\
  I3D + Kinetics & 12.3 & 6.6 & 10.7 & 9.7                    & 11.6 & 8.2 & 12.9 & 5.1 & 14.4 & 6.6 \\
  \hline
  $u$-net PIT \cite{yu2017permutation} & -- & 7.3 & 11.4 & {\bestcell 10.3}
  & -- & {\bestcell 8.8} & -- & 5.9 & -- & {\bestcell 8.1}\\ Deep Sep. \cite{huang2015separation} & -- & 1.3 & 3.0 & 8.7 & -- & 1.9 &
  -- & 0.8 & -- & 2.2 \\ \bottomrule \end{tabularx} \vspace{-2.5mm} \caption{{\small
  Source separation results on speech mixtures from the VoxCeleb
  (broken down by gender of speakers in mixture) and transfer to the
  simple GRID dataset.  We evaluate the on/off-screen sound prediction
  error (On/off) using $\ell_1$ distance to the true log-spectrograms (lower
  is better). We also use blind source separation metrics (higher is
  better) \cite{vincent2006performance}.}} \label{tab:long} }\end{table}

\begin{SCtable}[\sidecaptionrelwidth][h!]
\begin{minipage}[t!]{0.6\linewidth}
\vspace{-1.5mm}
\small
\centering
\begin{tabularx}{1.0\linewidth}{L{.4\linewidth} cccc}
\toprule
\multicolumn{5}{c}{VoxCeleb short videos (200ms)}\\
& On-SDR & SDR & SIR & SAR\\
\hline
Ours (on/off)                        & {\bestcell 7.6} & 5.3 & 7.8 & 10.8 \\
Hou \etal \cite{hou2017audio}        & 4.5       & -- & -- & -- \\
Gabbay \etal \cite{gabbay2017seeing} & 3.5       & -- & -- & -- \\
\hline
PIT-CNN \cite{yu2017permutation}     & --         & {\bestcell 7.0} & 10.1 & {\bestcell 11.2}\\
$u$-net PIT \cite{yu2017permutation} & --         & {\bestcell 7.0} & {\bestcell 10.3} & 11.0 \\
Deep Sep. \cite{huang2015separation} & --       &  2.7 & 4.2 & 10.3 \\
\bottomrule
\end{tabularx}
\end{minipage}\hfill

\begin{minipage}[t!]{0.4\linewidth}

\caption{{\small Comparison of audio-visual and audio-only separation
methods on short (200ms) videos. We compare SDR of the on-screen audio
prediction (On-SDR) with audio resampled to 2 kHz.}}\label{tab:short} 
\end{minipage}

\end{SCtable}

 We augment our audio-visual network with a $u$-net encoder-decoder
\cite{ronneberger2015u,isola2016image,michelsanti2017conditional} that
maps the mixture sound to its on- and off-screen components
(\fig{fig:sepnet}).  To provide the $u$-net with video information, we
include our multisensory network's features at three temporal scales:
we concatenate the last layer of each temporal scale with the layer of
the encoder that has the closest temporal sampling rate. Prior to
concatenation, we use linear interpolation to make the video features
match the audio sampling rate; we then mean-pool them spatially, and
tile them over the frequency domain, thereby reshaping our 3D CNN's
time/height/width shape to match the 2D encoder's time/frequency
shape.  We use parameters for $u$-net similar
to \cite{isola2016image}, adding one pair of convolution layers to
compensate for the large number of frequency channels in our
spectrograms. We predict both the magnitude of the log-spectrogram and
its phase (we scale the phase loss by 0.01 since it is less
perceptually important). To obtain waveforms, we invert the predicted
spectrogram. We emphasize that our model uses raw video, with no
preprocessing or labels (\eg no face detection or pretrained
supervised features).

{\setlength\textfloatsep{.0in}
\begin{figure}[t!]
  \includegraphics[width=1.0\linewidth]{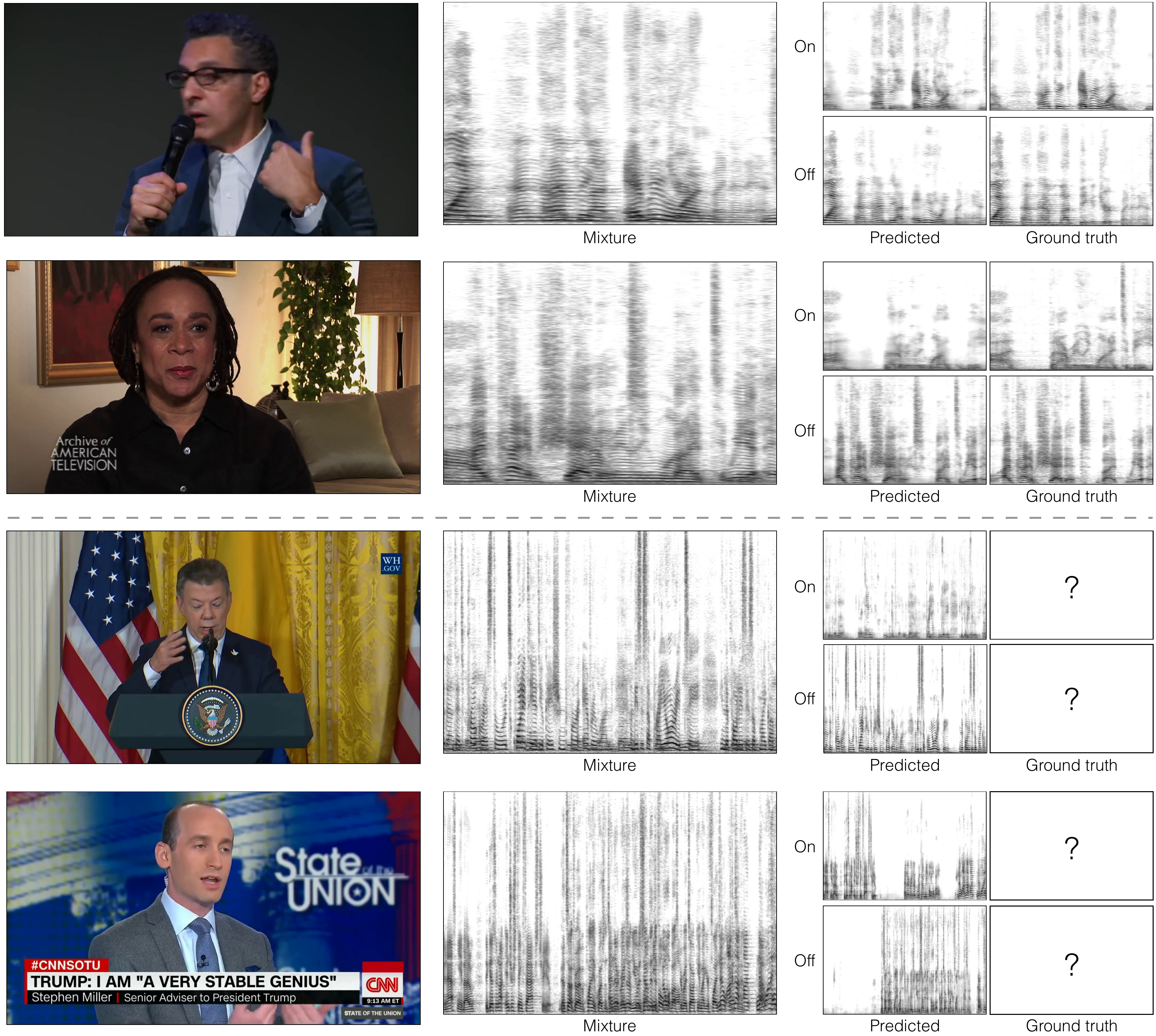}
  \caption{Qualitative results from our on/off-screen separation
    model.  We show an input frame and spectrogram for two synthetic
    mixtures from our test set, and two in-the-wild internet videos
    containing multiple speakers.  The first (a male/male mixture)
    contains more artifacts than the second (a female/male
    mixture). The third video is a real-world mixture in which a
    female speaker (simultaneously) translates a male Spanish speaker
    into English. Finally, we separate the speech of two (male)
    speakers on a television news show. Although there is no ground
    truth for these real-world examples, the source separation method
    qualitatively separates the two voices. Please refer to our
    webpage (\projecturl) for video source separation
    results. \label{fig:sepqual}}\vspace{-3mm}
\end{figure}}

\setcapspacing
 
\xpar{Training} We evaluated our model on the task of separating
speech sounds using the VoxCeleb dataset \cite{nagrani2017voxceleb}.
We split the training/test to have disjoint speaker identities
(72\%, 8\%, and 20\% for training, validation, and test). During
training, we sampled 2.1-sec. clips from longer 5-sec. clips, and
normalized each waveform's mean squared amplitude to a constant
value. We used spectrograms with a 64 ms frame length and a 16 ms step
size, producing $128 \times 1025$ spectrograms.  In each mini-batch of
the optimization, we randomly paired video clips, making one the
off-screen sound for the other. We jointly optimized our multisensory
network and the $u$-net model, initializing the weights using our
self-supervised representation (see supplementary material for details).

\xsub{Evaluation}

We compared our model to a variety of separation methods: 1) we
replaced our self-supervised video representation with other features,
2) compared to audio-only methods using blind separation methods, 3)
and compared to other audio-visual models.

\xpar{Ablations} Since one of our main goals is to evaluate the
quality of the learned features, we compared several variations of our
model (\tbl{tab:long}). First, we replaced the multisensory features
with the I3D network~\cite{carreira2017quo} pretrained on the Kinetics
dataset --- a 3D CNN-based representation that was very effective for
action recognition (\sect{sec:action}).  This model performed
significantly worse (11.4 vs. 12.3 spectrogram $\ell_1$ loss for
\eqn{eq:onoffloss}). One possible explanation is that our pretraining
task requires extensive motion analysis, whereas even single-frame
action recognition can still perform
well~\cite{simonyan2014two,karpathy2014large}.

We then asked how much of our representation's performance comes from
motion features, rather than from recognizing properties of the
speaker (\eg gender). To test this, we trained the model with only a
single frame (replicated temporally to make a video). We found a
significant drop in performance (11.4 vs. 14.8 loss). The drop was
particularly large for mixtures in which two speakers had the same
gender --- a case where lip motion is an important cue.

One might also ask whether early audio-visual fusion is helpful ---
the network, after all, fuses the modalities in the spectrogram
encoder-decoder as well. To test this, we ablated the audio stream of
our multisensory network and retrained the separation model. This
model obtained worse performance, suggesting the fused audio is
helpful even when it is available elsewhere. Finally, while the
encoder-decoder uses only monaural audio, our representation uses
stereo. To test whether it uses binaural cues, we converted all the
audio to mono and re-evaluated it. We found that this did not
significantly affect performance, which is perhaps due to the
difficulty of using stereo cues in in-the-wild internet videos (\eg
39\% of the audio tracks were mono). Finally, we also transferred
(without retraining) our learned models to the GRID dataset
\cite{cooke2006audio}, a lab-recorded dataset in which people speak
simple phrases in front of a plain background, finding a similar
relative ordering of the methods.

\xpar{Audio-only separation} To get a better understanding of our
model's effectiveness, we compared it to audio-only separation
methods. While these methods are not applicable to on/off-screen
separation, we modified our model to have it separate audio using an
extra permutation invariant loss (\eqn{eq:pit}) and then compared the
methods using blind separation metrics~\cite{vincent2006performance}:
signal-to-distortion ratio (SDR), signal-to-interference ratio (SIR),
and signal-to-artifacts ratio (SAR). For consistency across methods,
we resampled predicted waveforms to 16 kHz (the minimum used by all
methods), and used the mixture phase to invert our model's
spectrogram, rather than the predicted phase (which none of the others
predict).

We compared our model to PIT-CNN~\cite{yu2017permutation}. This model
uses a VGG-style \cite{simonyan2014very} CNN to predict two soft
separation masks via a fully connected layer. These maps are
multiplied by the input mixture to obtain the segmented streams. 
While this method worked well on short clips, we found it failed on
longer inputs (\eg obtaining 1.8 SDR in the experiment shown
in \tbl{tab:long}). To create a stronger PIT baseline, we therefore
created an audio-only version of our $u$-net model, optimizing the PIT
loss instead of our on/off-screen loss, \ie replacing the VGG-style
network and masks with $u$-net. We confirmed that this model obtains
similar performance on short sequences (\tbl{tab:short}), and found it
successfully trained on longer videos. Finally, we compared with a
pretrained separation model~\cite{huang2015separation}, which is based
on recurrent networks and trained on the TSP
dataset~\cite{kabal2002tsp}.

We found that our audio-visual model, when trained with a PIT loss,
outperformed all of these methods, except for on the SAR metric, where
the $u$-net PIT model was slightly better (which largely measures the
presence of artifacts in the generated waveform). In particular, our
model did significantly better than the audio-only methods when the
genders of the two speakers in the mixture were the same
(\tbl{tab:long}). Interestingly, we found that the audio-only methods
still performed better on blind separation metrics when transferring
to the lab-recorded GRID dataset, which we hypothesize is due to the
significant domain shift.

\xpar{Audio-visual separation} We compared to the
audio-visual separation model of Hou \etal~\cite{hou2017audio}. This
model was designed for enhancing the speech of a previously known
speaker, but we apply it to our task since it is the most closely
related prior method. We also evaluated the network of Gabbay
\etal~\cite{gabbay2017visual} (a concurrent approach to ours). We trained
these models using the same procedure as ours (\cite{gabbay2017visual}
used speaker identities to create hard mixtures; we instead assumed
speaker identities are unknown and mix randomly). Both models take
very short (5-frame) video inputs. Therefore, following
\cite{gabbay2017visual} we evaluated 200ms videos
(\tbl{tab:short}). For these baselines, we cropped the video around
the speaker's mouth using the Viola-Jones~\cite{viola2001rapid}
lip detector of \cite{gabbay2017visual} (we do not use face detection
for our own model). These methods use a small number of frequency
bands in their (Mel-) STFT representations, which limits their
quantitative performance. To address these limitations, we evaluated
only the on-screen audio, and downsampled the audio to a low, common
rate (2 kHz) before computing SDR. Our model significantly outperforms
these methods. Qualitatively, we observed that \cite{gabbay2017visual}
often smooths the input spectrogram, and we suspect its performance on
source separation metrics may be affected by the relatively small
number of frequency bands in its audio representation.

\xsub{Qualitative results}

Our quantitative results suggest that our model can successfully
separate on- and off-screen sounds. However, these metrics are limited
in their ability to convey the quality of the predicted sound (and are
sensitive to factors that may not be perceptually important, such as
the frequency representation). Therefore, we also provide qualitative
examples.

\xpar{Real mixtures} In \fig{fig:sepqual}, we show results for two
synthetic mixtures from our test set, and two real-world mixtures: a
simultaneous Spanish-to-English translation and a television interview
with concurrent speech. We exploit the fact that our model is fully
convolutional to apply it to these 8.3-sec. videos ($4 \times$ longer
than training videos). We include additional source separation
examples in the videos on our webpage. This includes a
random sample of (synthetically mixed) test videos, as well as results
on in-the-wild videos that contain both on- and off-screen sound.

\xpar{Multiple on-screen sound sources} To demonstrate our model's
ability to vary its prediction based on the speaker, we took a video
in which two people are speaking on a TV debate show, visually masked
one side of the screen (similar to \cite{fisher2000learning}), and ran
our source separation model.  As shown in \fig{fig:teaser}, when the
speaker on the left is hidden, we hear the speaker on the right, and
vice versa.  Please see our video for results.

\xpar{Large-scale training} We trained a larger variation of our model
on significantly more data. For this, we combined the VoxCeleb and
VoxCeleb2~\cite{chung2018voxceleb2} datasets (approx. 8$\times$ as
manys videos), as in~\cite{afouras2018conversation}, and modeled
ambient sounds by sampling background audio tracks from AudioSet
approximately 8\% of the time. To provide more temporal context, we
trained with 4.1-sec. videos (approx. 256 STFT time samples). We also
simplified the model by decreasing the spectrogram frame length to 40
ms (513 frequency samples) and increased the weight of the phase loss
to 0.2. Please see our webpage for results.

\vspace{-1.25mm}
\xsect{Discussion}

In this paper, we presented a method for learning a temporal
multisensory representation, and we showed through experiments that it
was useful for three downstream tasks: (a) pretraining action
recognition systems, (b) visualizing the locations of sound sources,
and (c) on/off-screen source separation. We see this work as opening
two potential directions for future research.  The first
is developing new methods for learning fused multisensory
representations.  We presented one method ---
detecting temporal misalignment --- but one could also incorporate other
learning signals, such as the information provided by ambient
sound~\cite{owens2016ambient}.  The other
direction is to use our representation for additional audio-visual
tasks. We presented several applications here, but there are other
audio-understanding tasks could potentially benefit from visual
information and, likewise, visual applications that could benefit from
fused audio information.

\vspace{-0.25mm}
\xpar{Acknowledgements} This work was supported, in part, by DARPA grant
FA8750-16-C-0166, U.C. Berkeley Center for
Long-Term Cybersecurity, and Berkeley DeepDrive. We thank Allan Jabri, David Fouhey, Andrew Liu, Morten
Kolb\ae k, Xiaolong Wang, and Jitendra Malik for the helpful discussions.

{\small
  \bibliographystyle{splncs}
  \bibliography{align} %
}

\renewcommand{\thesection}{A\arabic{section}}
\renewcommand{\thefigure}{A\arabic{figure}}
\setcounter{section}{0}
\setcounter{figure}{0}

\xsect{Optimization}
\label{sec:opt}
We use the following optimization procedures in our experiments:

\xpar{Temporal alignment} We trained the network from scratch with a
batch size of 15 (spread over 3 GPUs) for 650,000 iterations
(approximately 17 days). We used SGD with momentum, and decreased the
learning rate from its initial value of 0.01 by a factor of 2 every
200,000 iterations. In each mini-batch, we included both the video's
authentic and shifted audio, reusing the feature activations from the
video subnetwork between these examples for efficiency. To make
training easier, we pretrained the network at a lower video
frame-rate. For the first 30,000 iterations of training, we used a 7.5
Hz (a quarter the normal rate) and used a correspondingly larger batch
size of 45. We randomly initialized the network's weights using
\cite{he2015delving}.

\xpar{Action recognition} We fine-tuned the networks using a batch
size of 24 (spread over on 3 GPUs) for 30,000 iterations, dropping the
learning rate by a factor of 10 every 12,000 iterations. We used Adam
\cite{kingma2014adam} with an initial learning rate of $10^{-4}$. We
used data augmentation: cropping, left/right flipping, and shifting
the audio by a small amount (up to 1 frame).

\xpar{Source separation} We fine-tuned the networks for
160,000 iterations with a batch size of 18 videos (spread over 3
GPUs), dropping the learning rate by a factor of 10 after 120,000
iterations. We used Adam \cite{kingma2014adam} with an initial
learning rate of $10^{-4}$, and used cropping and left/right flipping
visual augmentations. We trained the model that operated on shorter
(200ms) sequences for 250,000 iterations total, and dropped the
learning rate after 150,000 iterations.

\xsect{Baseline model details}
\label{sec:appbaseline}
For the variation of our model that uses a spectrogram instead of a
waveform, we replaced the audio subnetwork with a 2D CNN. For this, we
adapted the ResNet-18 of \cite{he2016deep}, truncating it after the
{\tt conv3\_1} layer. We performed temporal subsampling after the {\tt
  conv2\_2} layers and frequency subsampling after the {\tt conv1} and
{\tt conv2\_2} and {\tt conv3\_1} layers. We used log-spectrograms
with a step size of 8 ms, a window length of 24 ms (corresponding to a
257 frequency bins). Before fusion, we applied a $3 \times F$
convolution with 512 channels, where $F$ is the number of frequency
channels. The subsequent layers of the network are the same as in
\fig{fig:network}. 

\begin{figure}%
  \centering
  \includegraphics[width=0.95\linewidth]{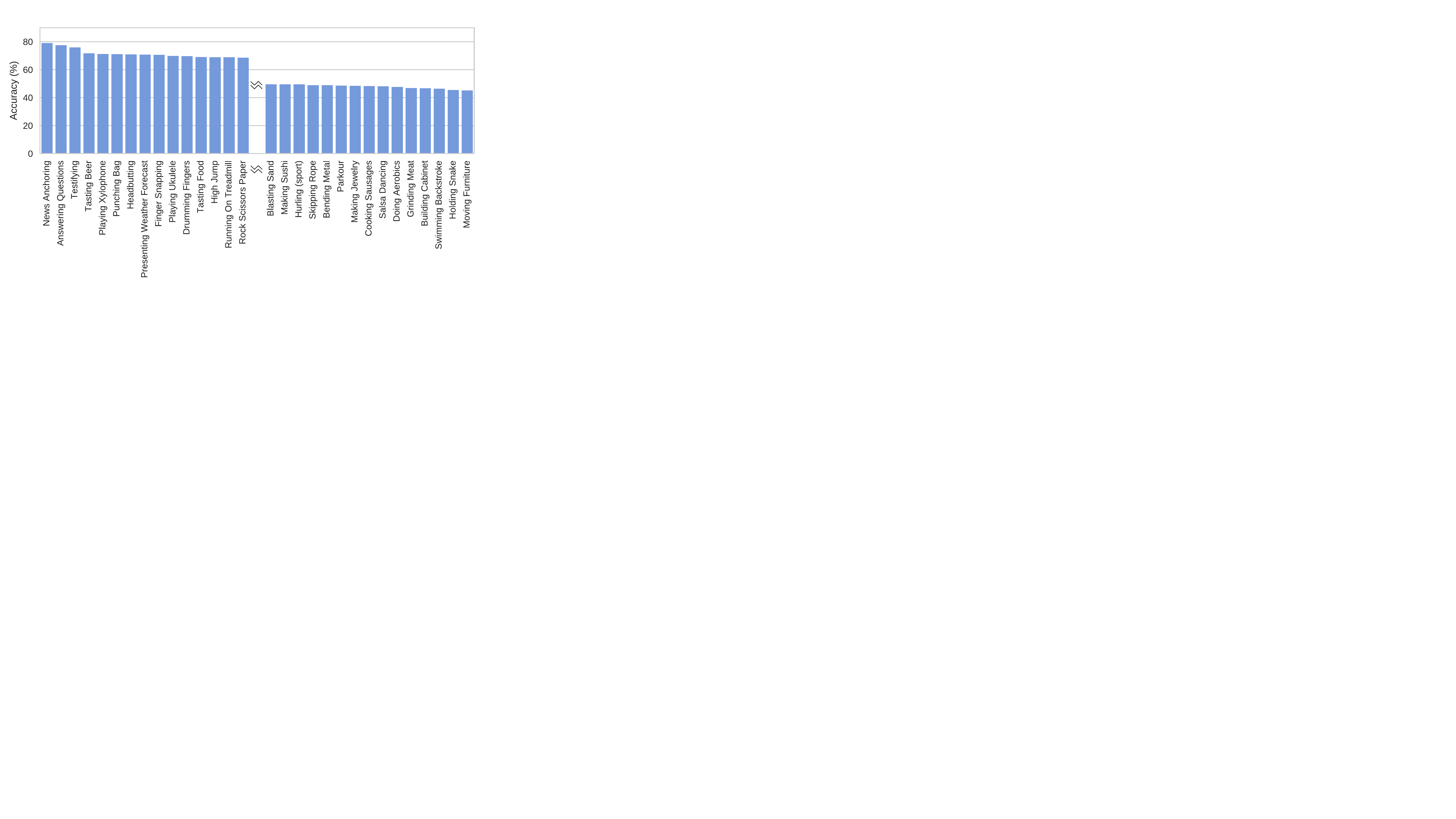}
  \vspace{-3mm}
  \caption{{\small Accuracy of our model in predicting audio-visual synchronization for the classes in the Kinetics dataset. Chance is 50\%.}}
  \label{fig:kineticsacc}
\end{figure}

\end{document}